\documentclass[letterpaper, 10 pt, conference]{ieeeconf}  

\IEEEoverridecommandlockouts                              

\usepackage{amsmath}  
\usepackage{amsfonts} 
\usepackage{bm}       

\usepackage{graphicx}

\usepackage{booktabs}
\usepackage{tabularx}
\usepackage{arydshln}

\title{\LARGE \bf
Steering Prediction via a Multi-Sensor System for Autonomous Racing
}

\author{Zhuyun Zhou$^{1,2}$, Zongwei Wu$^{3}$, Florian Bolli$^{2}$, Rémi Boutteau$^{4}$, Fan Yang$^{1}$, \\ Radu Timofte$^{3}$, Dominique Ginhac$^{1}$, and Tobi Delbruck$^{2}$
\thanks{This research is financed in part by the French National Research Agency through ANR CERBERE (ANR-21-CE22-0006), by the Swiss-European Mobility Program, and by the Alexander von Humboldt Foundation.}
\thanks{$^{1}$Zhuyun Zhou, Fan Yang, and Dominique Ginhac are with University of Burgundy, Dijon, France. {\tt\small \{Zhuyun\_Zhou@etu., fanyang@, dginhac@\}u-bourgogne.fr }}%
\thanks{$^{2}$Zhuyun Zhou, Florian Bolli, and Tobi Delbruck are with Sensors Group, Institute of Neuroinformatics, UZH/ETH Zürich, Switzerland. {\tt\small firstname.lastname@uzh.ch}}
\thanks{$^{3}$Zongwei Wu and Radu Timofte are with University of Wurzburg, Germany. {\tt\small firstname.lastname@uni-wuerzburg.de}}
\thanks{$^{4}$Rémi Boutteau is with Université  Rouen Normandie, INSA Rouen Normandie, Université  Le Havre Normandie, Normandie Université, LITIS UR 4108, Rouen, France. {\tt\small remi.boutteau@univ-rouen.fr}}
}

\begin{document}

\maketitle
\thispagestyle{empty}
\pagestyle{empty}

\begin{abstract}
Autonomous racing has rapidly gained research attention. Traditionally, racing cars rely on 2D LiDAR as their primary visual system. In this work, we explore the integration of an event camera with the existing system to provide enhanced temporal information. Our goal is to fuse the 2D LiDAR data with event data in an end-to-end learning framework for steering prediction, which is crucial for autonomous racing. To the best of our knowledge, this is the first study addressing this challenging research topic. We start by creating a multisensor dataset specifically for steering prediction. Using this dataset, we establish a benchmark by evaluating various SOTA fusion methods. Our observations reveal that existing methods often incur substantial computational costs. To address this, we apply low-rank techniques to propose a novel, efficient, and effective fusion design. We introduce a new fusion learning policy to guide the fusion process, enhancing robustness against misalignment. 
Our fusion architecture provides better steering prediction than LiDAR alone, significantly reducing the RMSE from 7.72 to 1.28. Compared to the second-best fusion method, our work represents only 11\% of the learnable parameters while achieving better accuracy. The source code, dataset, and benchmark will be released to promote future research.

\end{abstract}

\section{INTRODUCTION}

Autonomous racing has recently attracted growing research interest as it provides a novel platform for testing and evaluating emerging technologies in autonomous driving. Researchers can significantly reduce costs by utilizing scaled-down prototypes like the F1tenth, while still effectively testing hardware systems and software algorithms. 

In this work, we address the challenge of steering angle prediction, which is essential for keeping the vehicle within track boundaries. This challenge is particularly difficult in F1tenth prototypes, where most vehicles are equipped only with a 2D LiDAR for environment perception~\cite{evans2024unifying, babu2020f1tenth, evans2023bypassing}. As a result, \cite{snider2009automatic, becker2023model, lyu2022toward} have emerged among the most widely used approaches for steering control thanks to their advantages from known geometric properties of the vehicle. Nevertheless, these methods are based on the 2D LiDAR, which is the primary sensor for most F1tenth prototypes. Since the 2D LiDAR is only sensitive to depth changes, existing methods are limited by the lack of spatial awareness along the Y- and Z-axis of the ego F1tenth coordinate basis, as well as the temporal clues. When it comes to a scenario where the vehicle operates at high speeds in dynamic environments, these methods may suffer instability from perception delay, which can hinder the vehicle's ability to make rapid decisions. 

To overcome these limitations, we first enhance the conventional mono-sensor system by integrating an event camera to create a multisensor setup. As a bio-inspired device, the event camera \cite{gallego2020event} 
provides sparse output with high time resolution, high dynamic range, and energy efficiency. These features make it an ideal, though not yet fully explored, component for autonomous racing. By combining distance data from the 2D LiDAR with the event camera’s dynamic visual output, we aim to develop a multisensor fusion system that leverages the strengths of both sensors to enhance steering predictions.

\begin{figure}[t]
\centering

\includegraphics[width=.99\linewidth]{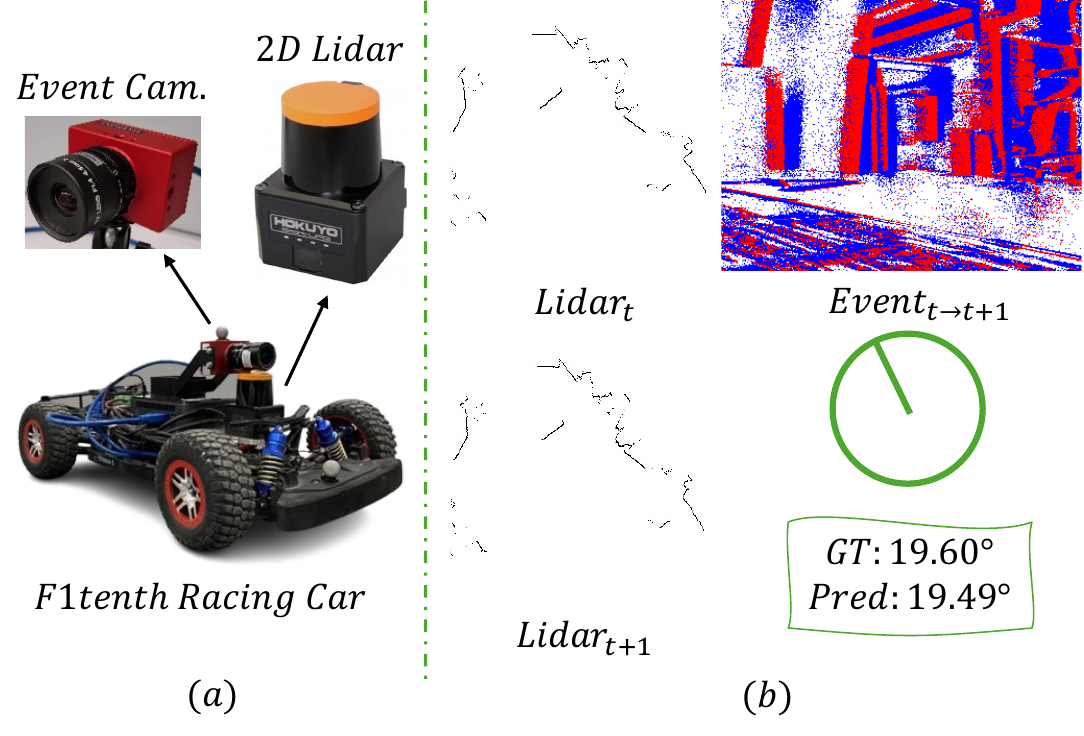}
\caption{
(a) The F1tenth racing car used in our experiments is equipped with a DAVIS346 event camera \cite{Brandli2014-davis} and a Hokuyo 2D LiDAR sensor.
(b) Our network processes two consecutive LiDAR scans captured at times \( t \) and \( t+1 \), along with an event-accumulated frame that includes all ``on" and ``off" brightness change events occurring between \( t \) and \( t+1 \). The network's objective is to predict the steering angle at time \( t+1 \). The LiDAR depth maps are depicted with a blank background, indicating areas with no data, while black pixels correspond to scan points, with the intensity of darkness reflecting proximity. We show that it is possible to leverage the joint benefit within such a multisensor system to achieve accurate steering prediction.
}
\label{Fig-car}
\vspace{-3mm}
\end{figure}

Secondly, for accurate and rapid steering prediction, we propose an end-to-end learning-based approach capable of fusing both sensory inputs. It is important to note that, unlike 3D dense point cloud data commonly used in existing methods, no viable learning-based approach currently exists for the primary 2D LiDAR data. In terms of event data, \cite{maqueda2018event} pioneered an event-based steering prediction method. However, these existing methods have yet to address how to adapt such techniques to a multisensor setting.

To fill this gap, we propose a novel multisensor fusion framework specifically designed for F1tenth. To the best of our knowledge, this is the first work to address the integration of a 2D LiDAR and event camera for autonomous racing. While our overall architecture follows a traditional learning-based pipeline, featuring sensor data extraction, fusion, and prediction, our primary contribution lies in devising an effective information fusion strategy. This strategy maximizes the joint entropy between the two sensor inputs, while taking practical concerns into consideration such as efficiency, heterogeneous data representation, and sensor misalignment.

\section{RELATED WORK}
\label{fusion}

\noindent \textbf{Autonomous Racing and F1tenth Prototype:}
Recent advancements in computer technology have significantly accelerated autonomous vehicle research, leading to the rise of autonomous racing competitions. These events have drawn substantial attention from researchers \cite{karaman2017project,rosolia2019learning,srinivasa2019mushr,o2020f1tenth,wu2019improved}, focusing on the design and real-time onboard computation of 1/10 scale prototypes, which are commonly used as proof-of-concept platforms in academic settings. Given the emphasis on energy efficiency, strict demands are placed not only on the onboard computational units but also on the choice of visual and sensor systems. For example, UC Berkeley’s BARC project utilizes ODROID single-board computers and custom wheel encoders to support its operations \cite{gonzales2016autonomous, rosolia2019learning}. MIT’s RACECAR project \cite{karaman2017project} employs NVIDIA’s Jetson TX1 system-on-module, along with VESC and a sensor suite, including a 2D LiDAR, stereo camera, and IMU.

\noindent \textbf{Event Camera and Fusion:}
Event cameras have garnered significant research interest due to their asynchronous processing, high dynamic range, and energy efficiency. When combined with conventional RGB inputs, these cameras show great potential to revolutionize traditional vision and robotic tasks, such as autonomous driving \cite{zhou2023rgb,zhou2023event}, depth estimation \cite{gehrig2021combining}, and semantic segmentation \cite{zhang2023cmx}. Early approaches to sensor fusion focused on simple feature concatenation to produce joint outputs \cite{cao2022neurograsp,tomy2022fusing}. Building on this, many researchers have applied spatial and channel attention mechanisms to refine sensor features either before or after merging the information \cite{cao2021fusion, zhou2023rgb}. More recent work \cite{zhang2023cmx,sun2022event,cao2024embracing} leverages transformer mechanisms for deeper feature modeling, improving robustness against misalignment but with increased computational demands.

Another research direction explores LiDAR-Event fusion, focusing on enhancing 3D dense point clouds with 2D event data. Despite the difference in target sensors, the underlying motivation is similar to RGB-Event fusion, as demonstrated by the effectiveness of transformer-based self and cross-attention mechanisms for feature fusion \cite{wan2023rpeflow,zhou2024bring,munir2023multimodal}. In practice, these fusion designs are quite similar to those used in RGB-Event fusion \cite{sun2022event,zhang2023cmx}, with only minor variations, hence can barely correspond to the real-time computational exigence for autonomous racing.

\noindent \textbf{Steering Angle Prediction:} Steering angle prediction is a crucial research topic for autonomous driving. There are many surveys in the literature \cite{gidado2020survey,saleem2021steering}. Here we only review some closely related learning-based works.

Learning-based approaches estimate steering angle predictions by directly mapping visual observations to control actions. This is typically achieved by designing a deep neural network that takes observations from two different time stamps and predicts the steering angle, supervised by GT targets \cite{hu2020learning,wu2021sdlv}. For example, mapping pixels from two RGB frames or mapping vovels from two LiDAR scan can produce a single steering prediction. However, both RGB and LiDAR are limited by a lack of temporal awareness.

Recently, with advances in event camera technology and the release of large-scale datasets like DDD17 \cite{binas2017ddd17} and DDD20 \cite{hu2020ddd20}, many studies \cite{munir2023multimodal,maqueda2018event} have explored jointly estimating the steering angle using event data or a combination of event data and other sensors. The latter, multisensor approaches (event + X) typically demonstrate better accuracy. However, directly adapting these existing methods to the F1tenth platform is not straightforward due to differences in available sensor systems and the platform's cost-efficiency requirements. In this work, we explore the feasibility of fusing 2D LiDAR with event camera data to achieve both high performance and efficiency in steering angle prediction for the F1tenth platform. To the best of our knowledge, this research topic is addressed for the first time in our setting.

\section{Sensor Setup and Dataset}

\subsection{Our F1tenth Prototype}

Figure \ref{Fig-car} shows the F1tenth car. This 1/10 scale vehicle, based on the Traxxas Slash RC platform, is a variant of the standard F1tenth car\footnote{https://f1tenth.org/build.html}. The car is equipped with an Intel NUC12 as its primary onboard computer running ROS noetic on Ubuntu20, a Hokuyo 2D LiDAR with a 10-meter range and 40Hz scanning frequency, a DAVIS346 event camera and a VESC MK6 electronic speed controller that manages the brushless motor and includes an integrated IMU.
The car's state estimation is based on an extended Kalman Filter, which integrates data from the motor encoder and IMU to generate odometry information. The drifting odometry is refined by the SLAM localization algorithm that determines a more accurate position of the car, by aligning LiDAR scans with the pre-mapped environment. 
Both mapping and localization were performed, using the ROS-Cartographer\footnote{https://google-cartographer-ros.readthedocs.io/en/latest/} SLAM algorithm. During the data collection, the car operated autonomously using the MPA algorithm, and all sensor data and system states were recorded and saved within the Rosbags for further analysis.

\subsection{ROS Messages and  Synchronization}
\label{sync}

Event cameras operate asynchronously, capturing per-pixel brightness changes in real time. The output is a continuous stream of events, where each event represents a change in brightness at a specific pixel location and time \cite{gallego2020event}. Formally, the event stream is defined as:
\begin{equation}
\varepsilon = \{ e_i | e_i = ((x_i, y_i), t_i, p_i), t_i \in [t_{start}, t_{end}] \}
\label{Eq-event}
\end{equation}
where $ e $ denotes an individual event, $ (x, y) $ represents the pixel coordinates, $ t $ is the timestamp of the event, and $ p $ is the 1-bit polarity with $ p \in \{ +1, -1 \} $, indicating whether the brightness at that pixel increases (+$1$) or decreases (-$1$).

Given that the 2D LiDAR operates at a frequency of 40Hz, and to facilitate multi-modal fusion, we accumulate the ``on" and ``off" events from the event camera according to the timestamps of the LiDAR scans. As a result, each pair of consecutive LiDAR scans is matched with a single accumulated event frame. 

Steering angle extraction follows a similar approach. Given the imperfect synchronization of ROS messages, we associate each LiDAR scan with the steering angle whose timestamp is closest to that of the LiDAR scan, doing our best to ensure temporal alignment across all modalities.

\subsection{Cross-Sensor Calibration and Projection}

As suggested in \cite{zhou2024bring}, when 3D LiDAR data is projected into 2D, or the event camera's perspective, LiDAR and event data share a similar structural manifold. Building on this insight, we propose projecting our 2D LiDAR observations into the 2D event camera view. This approach not only brings the sensor representations closer together but also simplifies LiDAR processing, making it more efficient and lightweight in 2D compared to 3D.

To achieve such a transformation, we base on the hypothesis of the pinhole camera model. The projection model can be described as follows:
\begin{equation}
\label{3d2d}
\mathbf{p}_\text{image} = K \left[ R | t \right] \mathbf{P}_\text{LiDAR} \\
\end{equation}
where $ \mathbf{P}_\text{LiDAR} $ is the 3D point in the LiDAR's coordinate system; 
$ R \in \mathbb{R}^{3 \times 3} $ is the rotation matrix that accounts for the relative orientation between the LiDAR and the camera; 
$ t \in \mathbb{R}^{3 \times 1} $ is the translation vector that represents the relative position of the LiDAR with respect to the camera; 
$ K \in \mathbb{R}^{3 \times 3} $ is the intrinsic matrix of the event camera, which encodes the camera’s focal length and principal point.

It is worth noting that the rotation and translation (\(\left[ R | t \right]\)) values are assigned using approximate estimates, which is a common practice and standard configuration on the F1tenth vehicle utilizing the ROS platform. These approximations inherently introduce a certain degree of error. Therefore, it can impact the performance and robustness during multi-sensor fusion. Therefore, we have designed a novel, effective, yet efficient method to take this misalignment issue into consideration during the learning phase.

\subsection{Cleansing and Final Version of the Dataset}

To ensure data consistency, we remove the initial and final portions of each ROS bag that contain preparatory or irrelevant messages. After this cleansing process, we retain a total of 27452 valid event-LiDAR-steering angle pairs. For the experiments, we utilize 21576 pairs from five distinct ROS bags for training, while 5876 pairs from two other separate ROS bags are reserved for testing. This ensures a robust evaluation of our model on diverse yet representative data subsets. Our dataset will be made publicly available to foster and support future research in this domain.

\begin{figure*}[t]
\centering
\includegraphics[width=.99\linewidth]{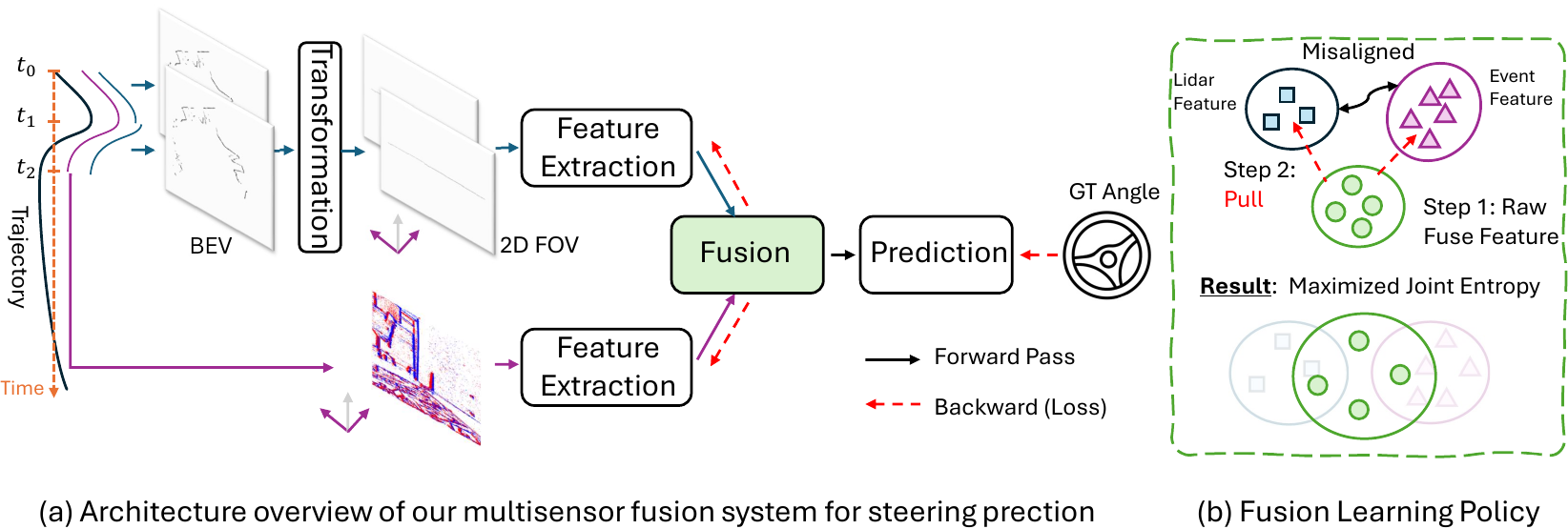}
\caption{
(a) Architecture Overview. \textbf{Please, zoom in} for better visualization of the BEV and 2D FOV view of the 2D LiDAR point. It is important to note that, unlike 3D LiDAR, the 2D LiDAR points form only a quasi-vertical line after projection. This characteristic makes our multisensor fusion particularly challenging, and, to the best of our knowledge, this specific issue is being addressed for the first time. The overall architecture follows a conventional feature extraction, fusion, and decoding pipeline. However, we introduce a novel fusion method with a new learning policy, as illustrated in (b), to fully exploit the mutual benefits between the 2D LiDAR and event camera, resulting in an efficient yet effective fusion strategy that maximizes joint entropy.
}
\label{Fig-archi}
\vspace{-3mm}
\end{figure*}

\section{Our Method}

Figure \ref{Fig-archi} illustrates the overall architecture of our end-to-end learning approach. The network takes as input two consecutive 2D LiDAR scans, along with the event stream from the corresponding time period. We implement a dual-feature extraction pipeline, processing the LiDAR scans and event streams in parallel. To enhance processing speed, we fuse the two LiDAR scans at the input stage, allowing us to use a single encoder for LiDAR feature extraction. After extracting the features from both the LiDAR scans and the event stream, we apply a feature fusion method to generate a joint feature representation. This fused feature is then passed through a decoder to estimate the steering angle. The entire network is trained in a fully supervised manner, using GT steering angles. Additionally, we incorporate internal supervision during the feature fusion stage for further refinement. Technical details of each step are described as follows.

\subsection{Feature extraction}

Without loss of generality and for the ease of review, we denote two consecutive LiDAR scans as $L_{t_1}$ and $L_{t_2}$, representing the scans acquired at timestamp $t_1$ and $t_2$, respectively. Following Eq. \ref{3d2d}, we project the LiDAR scans and obtain the associated 2D representations or the 2D depth maps, denoted as  $D_{t_1}$ and $D_{t_2}$. Note that each depth map only has 1 channel, referring to the geometric distance along the X-axis of the ego vehicle basis. These depth map are then concatenated along the channel dimension to form the 2-channel depth input $D$.

For the event camera, we follow the process outlined in Section \ref{sync} to obtain the temporally-aligned event stream. ``Temporally-aligned" refers to extracting the event stream based on the LiDAR timestamps. Following prior works \cite{sun2022event, zhou2023event}, the event data is accumulated to form a 2D image, denoted as $E$.

For feature extraction, we choose the EfficientNet \cite{mingxing2019efficientnet} as the backbone thanks to its efficiency and effectiveness. We process LiDAR and event input in a parallel manner and obtain LiDAR/depth feature $f_D$ and event feature $f_E$, respectively. 

\subsection{Fusion Learning Policy}

Once we obtain the multimodal features $f_D$ and $f_E$, the next step is to create a shared output for joint decoding. Conventional fusion methods often rely on implicit fusion, which does not explicitly control the fusion process. These methods assume that both features are perfectly aligned and focus only on exploring their joint benefits. However, in our case, and also as indicated by previous research \cite{bai2022transfusion}, cross-sensor calibration is always imperfect, leading to misalignment between the two features.

To enhance robustness to misalignment, we propose a novel fusion learning policy that explicitly controls the fusion process. Given that $f_D$ and $f_E$ are inherently different and misaligned due to calibration errors, we model these differences as distances between the LiDAR and event domains. Thus, as shown in Figure \ref{Fig-archi}(b), the fusion task is simplified to finding a point that minimizes the distance to both domains.
  
Specifically, let $f_S$ be the shared/fused output of $f_D$ and $f_E$.  We use similarity loss to evaluate the pairings ($f_S$ - $f_D$) and ($f_S$ - $f_E$). In our implementation, we chose the Kullback-Leibler divergence ($KL$) to measure the feature similarity:
\begin{equation}
\label{kl}
\begin{split}
&\mathcal{L}_{div} = \mathcal{L}_{KL}(f_{S}, f_{D}) + \mathcal{L}_{KL}(f_{S}, f_{E}), \\
&\mathcal{L}_{KL}(A, B) = KL(A || B) + KL(B || A).
\end{split}
\end{equation}

Through iterative training, this approach encourages the fused feature $f_S$ to move in directions that reduce the domain distance, ideally, the middle point of these two domains, as shown in Figure \ref{Fig-archi}(b). Such a method is expected to in turn to better align both domain and effectively bridge the gap between geometric knowledge and event clues.

\begin{figure*}[t]
\centering
\includegraphics[width=.99\linewidth]{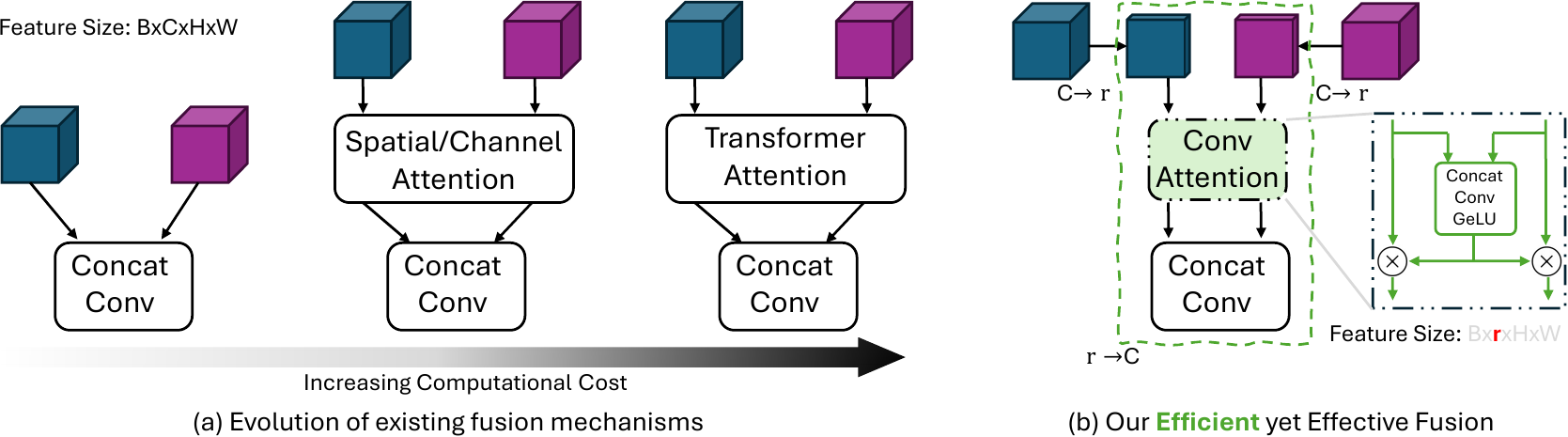}
\caption{
Comparison against existing fusion methods. Our approach begins by projecting the input features into a lower-dimensional latent space to reduce complexity. We then introduce a novel attention mechanism based on gated convolution. This combination makes our fusion technique both computationally efficient and highly effective,  surpassing existing (transformer) attention methods.
}
\label{Fig-fusion}
\vspace{-3mm}
\end{figure*}

\subsection{Efficient yet Effective Fusion}
The LiDAR features capture geometric changes along the X-axis of the ego F1tenth coordinate basis during the target time period, while the event features represent pixel-wise 2D image changes, which refers to changes along the Y- and Z-axes. Since ego-motion will lead to changes in any of these 3D directions, steering angle prediction can be simplified as mapping the feature differences between observations at two consecutive timestamps from any direction. In other words, a LiDAR-only or event-only visual system can thus meet the basic requirements for estimating the steering angle using visual cues. However, such predictions are based on incomplete information, focusing solely on either the X-axis or the Y- and Z-axes, without providing a comprehensive understanding of the ego motion.

A straightforward way to address this limitation is through late fusion, where the network generates two separate steering predictions, one based on LiDAR and one based on the event data, and computes their mean as the final output. The advantages and disadvantages of this approach are also straightforward. On the positive side, late fusion avoids the challenges of reconciling the heterogeneous representations of LiDAR and event data, as it operates in the shared space of steering angle prediction. Since the fusion happens at the output level, the predictions are naturally homogeneous or ``aligned". However, this approach also has drawbacks. As shown in Table \ref{Tab-Fusion}(Output fusion), first, it doubles the computational cost. Second, it fails to fully exploit the joint benefits of multimodal features during the deep feature modeling stage.

The fundamental principle of multimodal fusion is that different features provide heterogeneous yet complementary information for the target task. Various fusion methods have been proposed in the literature, as outlined in Section \ref{fusion}. Figure \ref{Fig-fusion}(a) offers a detailed overview of these approaches. It is evident that recent methods have become increasingly sophisticated, with more complex feature modeling designs aimed at preserving the most informative and unique aspects of each feature for more effective fusion. While these advanced methods can achieve impressive performance, they often involve substantial computational demands. This can be problematic for embedded systems with limited computational resources, as shown in Table \ref{Tab-Fusion}.

To address this challenge, we draw inspiration from recent advancements in low-rank techniques. We propose an efficient yet effective fusion method that projects input features into a low-dimensional latent space, approximating the low-rank structure of the feature map. By performing all feature manipulations in this low-rank space, we significantly reduce the number of learning parameters, making our approach more suitable for resource-constrained embedded systems.

As illustrated in Figure \ref{Fig-fusion}(b), we first project the LiDAR feature $f_D$ and the event feature $f_E$ into a low-rank space $r$, resulting in $r_D$ and $r_E$, respectively. This projection is achieved using simple $1 \times 1$ convolutions:
\begin{equation}
    r_{D} = Conv_{1\times1}(f_D); \quad    r_{E} = Conv_{1\times1}(f_E).
\end{equation}

Next, these features are combined to compute a joint attention map. Unlike traditional transformer-based attention mechanisms, our attention map is generated using gated convolution for improved efficiency:
\begin{equation}
    attn = GeLU(Conv_{1\times1}(Concat(r_D, r_E)))
\end{equation}

This joint attention map is then applied to both $r_D$ and $r_E$ to feature recalibration. 
\begin{equation}
    r_D' = attn \times r_D; \quad r_E' = attn\times r_E
\end{equation}

Finally, the refined features are merged and projected back into the original latent space to form the shared output $f_S$.

\subsection{Objective Function}
\label{loss}
Our network is end-to-end learnable, with full supervision from the GT steering angle. Such a supervision is achieved through a simple L2 loss. Therefore, in addition to the Equation \ref{kl}, our full loss function becomes:
\begin{equation}
    \mathcal{L} = \lambda \cdot \mathcal{L}_{div} + \mathcal{L}_{2}
\end{equation}
where $\lambda$ is a hyperparameter set to be 0.25.

\section{EXPERIMENTS}

\begin{table*}[t]
\footnotesize
\centering
\caption{
\textbf{Quantitative Comparison} with SOTA Event Fusion Approaches on our dataset. 
Metrics are Root Mean Squared Error (RMSE), Mean Absolute Error (MAE) Scores, and EVA (explained variance). 
Best values in bold.
}
\label{Tab-Fusion}
\begin{tabular}{l|c|c|c|c|c|c|c}
\toprule
Fusion Module & Pub. \& Year & Fusion Type & Params (M, $\downarrow$) & FLOPs (G, $\downarrow$) & RMSE ($\downarrow$) & MAE ($\downarrow$) & EVA ($\uparrow$) \\
\midrule
Event only & / & / & 4.828 & 1.582 & 7.164 & 4.737 & 0.190 \\

LiDAR only & / & / & 4.828 & 1.569 & 7.716 & 6.343 & 0.166 \\

Output fusion & / & mean & 2 $\times$ 4.828 &  3.151 & 5.980 & 4.588 & 0.462 \\
\hdashline

RAMNet \cite{gehrig2021combining} & RAL'21 & conv. & 500.363 & 4.132 & 2.292 & 1.730 & 0.918 \\

FPN-Fusion \cite{tomy2022fusing} & ICRA'22 & conv. & 38.328 & 3.208 & 1.744 & 1.290 & 0.952  \\

EFNet \cite{sun2022event} & ECCV'22 & transf. & 21.949 & 3.175 & 8.665 & 7.395 & 0.003 \\

DRFuser \cite{munir2023multimodal} & EAAI'23 & transf. & 18.666 & 3.272 & 4.562 & 3.682 & 0.773 \\

CMX \cite{zhang2023cmx} & TITS'23 & transf. & 79.331 & 3.290 & 1.590 & 1.156 & 0.962 \\

RENet \cite{zhou2023rgb} & ICRA'23 & attn & 71.521 & 3.275 & 2.971 & 2.377 & 0.897 \\

SAFusion \cite{jiang2024complementing} & CVPR'24 & attn & 155.616 & 3.442 & 2.504 & 1.950 & 0.902 \\
\midrule
\textbf{Ours} & - & conv. & \textbf{8.919} & \textbf{3.149} & \textbf{1.282} & \textbf{1.007} & \textbf{0.975} \\
\bottomrule
\end{tabular}
\vspace{-3mm}
\end{table*}

\subsection{Implementation Details}

We employ EfficientNet-B0 \cite{mingxing2019efficientnet} as the encoder to extract features from two consecutive LiDAR frames and one event frame. Standard data augmentation techniques, such as random flipping, are applied during training. The AdamW optimizer is used with an initial learning rate of \(1 \times 10^{-3}\), and a weight decay of \(1 \times 10^{-2}\). For learning rate scheduling, we utilize CosineAnnealingWarmRestarts, with a restart interval of 30 epochs. The network is implemented in PyTorch and trained for 200 epochs. The total training time is approximately 4.5 hours on an NVIDIA 4090 GPU.

\subsection{Metrics}
To quantitatively assess the performance of our model, we evaluate it using three standard regression metrics for steering angle prediction \cite{hu2020ddd20, maqueda2018event, munir2023multimodal}: Root Mean Squared Error (RMSE), Mean Absolute Error (MAE), and Explained Variance (EVA). These metrics provide a comprehensive understanding of both the precision and variance of our predictions relative to the ground truth.

\subsubsection{Root Mean Squared Error (RMSE)}
The RMSE is a widely used metric to measure the average magnitude of the prediction errors, providing insight into the accuracy of the model. It is computed as the square root of the mean squared differences between the predicted values \(\hat{y}_i\) and the ground truth values \(y_i\), as defined by the following equation:
\begin{equation}
\text{RMSE} = \sqrt{\frac{1}{N} \sum_{i=1}^{N} (y_i - \hat{y}_i)^2}
\end{equation}
where \(N\) is the total number of samples. RMSE penalizes larger errors more severely than smaller ones, making it sensitive to outliers. A lower RMSE indicates better model performance.

\subsubsection{Mean Absolute Error (MAE)}
MAE evaluates the average magnitude of errors in a set of predictions, without considering their direction. It is defined as:
\begin{equation}
\text{MAE} = \frac{1}{N} \sum_{i=1}^{N} |y_i - \hat{y}_i|
\end{equation}
Unlike RMSE, MAE treats all errors equally by taking the absolute difference between the ground truth and predictions. A lower MAE reflects higher prediction accuracy and robustness against outliers compared to RMSE.

\subsubsection{Explained Variance (EVA)}
The EVA measures the proportion of variance in the ground truth that is captured by the model. It is computed as:
\begin{equation}
\text{EVA} = 1 - \frac{\text{Var}(y - \hat{y})}{\text{Var}(y)}
\end{equation}
where \(\text{Var}(y)\) is the variance of the ground truth values and \(\text{Var}(y - \hat{y})\) is the variance of the residuals. An EVA of 1 signifies perfect prediction, while values close to 0 indicate that the model performs no better than predicting the mean of the ground truth. Negative values imply that the model is worse than a naive mean-based predictor.

\subsection{Benchmark and Comparison}

To assess the effectiveness of our approach, we compare it against 7 state-of-the-art event fusion methods, as summarized in Table \ref{Tab-Fusion}. These methods all fall into the categories presented in Figure \ref{Fig-fusion}, ie., conv-based, spatial/channel attention-based, or transformer attention-based. Note that the plain methods of these counterparts are not officially designed for steering angle. Hence, we only take the fusion methods and replace our fusion block with it. For fair comparison, all the methods are evaluated with the same loss function with both divergence and L2 losses, as mentioned in Section \ref{loss}.

It can be seen that (transformer) attention methods often surpass convolutional-based counterparts, but also come with additional learnable parameters and FLOPS. Differently, our convolution-based gated attention outperforms all the SOTA fusion alternatives, being more efficient while setting new SOTA records simultaneously. Our model can run at around 260 FPS. We plan to further test our model on the embedded onboard system in \textit{future work}.

\subsection{Ablation Studies}

We first conduct ablation studies on various choices of the latent space \( r \). The latent space is crucial for achieving efficient and effective fusion by significantly reducing computational costs. Table \ref{Tab-Latent} shows that extremely small latent spaces result in suboptimal performance, because the network cannot encode important features from the input data. We find that \( r = 16 \) delivers the best accuracy. Increasing \( r \) to 32 leads to a deterioration in accuracy, which may be attributed to the added complexity of learning in a higher-dimensional space. Despite this, our network shows resilience to different choices of \( r \), as most variants of \( r \) outperform the state-of-the-art solutions listed in Table \ref{Tab-Fusion}.

We also evaluate the key components of our network. As shown in Table \ref{Tab-Components}, replacing our EfficientNet backbone with a ResNet backbone results in a slight deterioration in accuracy (see \#1-\#2). This decline may be due to the higher number of learning parameters in ResNet compared to EfficientNet, which introduces additional complexity into the learning process, particularly for relatively simple LiDAR inputs. Additionally, we observe a significant accuracy improvement with the inclusion of our KL loss.

\begin{table}[t]
\footnotesize
\centering
\caption{
\textbf{Ablation Study on Latent Space}. Our choice is in bold.
}
\label{Tab-Latent}
\begin{tabular}{c|c|c|c}
\toprule
Latent Space & RMSE ($\downarrow$) & MAE ($\downarrow$) & EVA ($\uparrow$) \\
\midrule
2 & 1.949 & 1.516 & 0.951 \\

4 & 1.682 & 1.252 & 0.958 \\

8 & 1.389 & 1.079 & 0.970 \\

\textbf{16} & \textbf{1.282} & \textbf{1.007} & \textbf{0.975} \\

32 & 1.393 & 1.090 & 0.970 \\

\bottomrule
\end{tabular}
\end{table}

\begin{table}[t]
\footnotesize
\centering
\caption{
\textbf{Ablation study on key components.} L-B stands for the Lighter Backbone, i.e., EfficientNet.
}
\label{Tab-Components}
\begin{tabular}{c c c c c c c}
\toprule
\# & L-B & KL-loss & fus. & RMSE ($\downarrow$) & MAE ($\downarrow$) & EVA ($\uparrow$) \\
\midrule
1 & & & & 6.796 & 4.874 & 0.319 \\

2 & \checkmark & & & 6.257 & 4.645 & 0.400 \\

3 & \checkmark & \checkmark & & 3.542 & 2.891 & 0.887 \\

4 & \checkmark & \checkmark & \checkmark &\textbf{1.282} & \textbf{1.007} & \textbf{0.975} \\

\bottomrule
\end{tabular}
\vspace{-3mm}
\end{table}

\section{CONCLUSIONS}
This work introduces a novel multisensor fusion framework for steering angle prediction in autonomous racing, combining 2D LiDAR and event camera data. Our approach enhances steering prediction by integrating the high temporal resolution of event cameras with the distance accuracy of 2D LiDAR. We propose an end-to-end learning-based system that effectively fuses these sensor inputs, addressing sensor misalignment and data heterogeneity. Our framework shows significant improvements in steering accuracy and efficiency compared to existing methods. By providing our source code, dataset, and benchmark, we aim to facilitate further advancements in autonomous racing technology.


\noindent \textbf{Acknowledgment:} The authors would like to sincerely thank Liam Boyle, Nicolas Baumann, and Niklas Bastuck from PBL, ETH Zürich, for helping recordings and expertise of the F1tenth car, which are essential for this work.


\bibliographystyle{IEEEtran}
\bibliography{IEEEexample}

\end{document}